%% file: activation_power.tex
\documentclass[,dvipsnames]{article} 
\usepackage{iclr2020_conference,times}

\input{math_commands.tex}

\usepackage[hidelinks]{hyperref}
\usepackage{url}

\usepackage{courier}

\usepackage{listings}
\usepackage{graphicx}
\usepackage{caption}
\usepackage{subcaption}

\title{Power Consumption Variation over Activation Functions}

\author{Leon Derczynski\\
Department of Computer Science\\
IT University of Copenhagen\\
2300 Denmark \\
\texttt{leod@itu.dk} \\
}

\iclrfinalcopy 
\begin{document}

\maketitle

\begin{abstract}
The power machine learning models consume when making predictions can be affected by a model's architecture. This paper presents various estimates of power consumption for a range of different activation functions, a core factor in neural network model architecture design. Substantial differences in hardware performance exist between activation functions. This difference informs how power consumption in machine learning models can be reduced.
\end{abstract}

\section{Introduction}
The field of deep neural networks has reported strong progress in many problem areas, including natural language processing (NLP), image recognition, and game playing. Many of the advances in these areas have been the fruit of using larger and thus more computationally demanding neural networks. \cite{amodei18} find that the cost of training doubled every few months between the releases of AlexNet~\citep{krizhevsky2012imagenet} and AlphaZero~\cite{silver2018general}. In NLP, power consumption has also risen: \cite{strubell2019energy} determine the carbon footprint of a contemporary machine translation architecure search to be in the order of hundreds of intercontinental flights, for models that offer only marginal performance improvement.  



This paper examines activation functions, a core part of neural networks. The activation function is the non-linearity at the core of each network node. 
It is applied over the input and bias parameters at a given node for each inference that a model makes. This makes for a potentially large number of computation being required  to make predictions, predicated on network structure and size. When it comes to individual calculations, there is also broad variance. The complexity of low-level instructions for each these functions also varies widely, from the simple rectified linear unit to the transcendental hyperbolic tangent. This variance has the potential to lead to differences in power consumption. 

\section{Background}
The constraints on choice of a neural network activation function are (a) that it must be differentiable, and (b) that it must  have a continuous domain. These are required in order to train networks through backpropagation. A large range of functions fit these constraints, and as such, currently popular machine learning frameworks implement a broad range of activation functions.

The machine code for executing these functions can vary in complexity significantly. Figure~\ref{fig:listing} compares toy x86 code for rectified linear unit activation with code for a hyperbolic tangent (tanh) function. The tanh code is not only more complex, but also requires use of special resources such as an FPU (floating point unit). In practice, these functions are often run in a SIMD/SIMT structure, where a single instruction is performed over many data points at a time. However, x86 and CUDA SIMD instruction sets have similar restrictions: there is no direct tanh function for either, and instead a sequence of secondary operations have to be performed. While inference requires many other operations beyond calculating activation functions, the difference in scale of these functions' computation still leaves room for optimisation.

A further bottleneck is presented by hardware structure. The operations needed to compose some activation functions can require special hardware. This hardware can be scarce and therefore highly contended. For example, an NVIDIA V100 card's streaming multiprocessor (SM) has just one single special function unit (SFU) to sixteen 32-bit floating point cores, eight 64-bit floating point cores, and sixteen 32-bit integer cores~\citep{durant2017inside,markidis2018nvidia}. When computing an activation function requires special hardware, such as an SFU, the rest of the hardware unit (e.g. a CUDA warp) may be left idle until computation completes. Due to this bottleneck, use of these activation functions could lead to both increased power consumption (through fixed overheads incurred in the background as threads wait) and also slow models.

\begin{figure}
    \centering
    \begin{subfigure}[t]{0.45\textwidth}
        \centering
        \begin{lstlisting}
relu: push eax
      rol eax, 1
      <@\textcolor{gray}{xor eax, eax}@>
      <@\textcolor{gray}{and eax, 1}@>
      <@\textcolor{gray}{pop ebx}@>
      <@\textcolor{gray}{imul eax, ebx}@>
      ret
        \end{lstlisting}
        \caption{ReLU in x86-like code, with EAX holding a 32-bit float on entry. No floating point stack required; the function is applied bitwise with no branching. \textcolor{gray}{Grey} instructions take one micro-op. Timings from~\cite{agner}.}
    \end{subfigure}\hfill
    \begin{subfigure}[t]{0.45\textwidth}
        \centering
        \begin{lstlisting}
tanh: fst dword [tmp1]
      call exp
      fst dword [tmp2]
      <@\textcolor{gray}{fld dword [tmp1]}@>
      <@\textcolor{gray}{fchs}@>
      call exp
      fst dword [tmp1]
      <@\textcolor{gray}{fld dword [tmp2]}@>
      <@\textcolor{gray}{fsubr}@>
      <@\textcolor{gray}{fld dword [tmp2]}@>
      <@\textcolor{gray}{fld dword [tmp1]}@>
      <@\textcolor{gray}{fadd}@>
      <@\textcolor{gray}{fdiv}@>
      ret
exp:  fldl2e
      fmulp st1,st0
      fld1
      <@\textcolor{red}{fscale}@>
      fxch
      fld1
      fxch
      <@\textcolor{red}{fprem}@>
      <@\textcolor{red}{f2xm1}@>
      faddp st1,st0
      fmulp st1,st0
      ret

        \end{lstlisting}
        \caption{tanh in x86-like code; floating-point operations here begin '{\tt f}', which need FPUs and have higher execution times. \textcolor{red}{Red} instructions take more than ten micro-ops.}
    \end{subfigure}
    \caption{x86 style versions of ReLU vs. tanh.}
    \label{fig:listing}
\end{figure}

Current research increasingly addresses the energy impact of machine learning~\citep{Schwartz2019GreenA,sustainlp2020}. Large amounts of work has been done on reducing the training time of machine learning models~\citep{girosi1995regularization,prechelt1998automatic}; on reducing precision to afford bandwidth~\citep{woodland1989weight,wang2018training}; and increasing efficiency through parameter reduction~\citep{kim2016sequence,alvarez2019end}. Toolkits for measuring emissions impact have started to appear with particular detailed results in some geographic regions, that work on a limited range of hardware~\citep{henderson2020towards}. However, the specific impact that activation function choice has on power consumption has not been previously investigated.

\section{Experimental Setup}
\label{sec:setup}

The experiment goal is to gauge the power consumption impact of varying activation function type in a neural network. Although activation function choice can impact the length of the training phase, predicated on both architecture and training data, the resources needed to label one instance at inference time are predicated only on architecture. Thus, experiments do not need to consider training data in order to estimate impacts on inference-time power consumption.


\paragraph{Activation functions} Implementations of AlphaDropout,
CELU,
Dropout,
Dropout2d,
Dropout3d,
ELU,
GELU,
Hardshrink,
Hardtanh,
Identity,
LeakyReLU,
LogSigmoid,
LogSoftmax,
PReLU,
ReLU,
ReLU6,
RReLU,
SELU,
Sigmoid,
Softmax,
Softmin,
Softplus,
Softshrink,
Softsign,
Tanh,
 and Tanhshrink
in PyTorch~1.5.0~\citep{paszke2019pytorch} are evaluated. 

\paragraph{Network architecture} The test neural network had a 64-unit linear input layer, four hidden layers of 1024 units, and a sixteen unit linear output layer. The activation functions in the 4096 hidden nodes were varied as an experiment parameter. Each model was trained for each activation function for 2000 epochs using random data,  randomly initialised weights, and Adam optimisation.

\paragraph{Evaluation} The metric is power use, measured for different activation functions. Power consumption was proxied through wall clock time via Python's {\tt \small time.perf\_counter()}. This metric has known deficiencies: while a process consumes time it may impose a range of power overheads. Nevertheless, we expect that the specific variation in real power consumption is low enough across these similar experimental workloads, and spurious loads will be cushioned through the use of multiple runs. Experiments were run three times and means taken. Experiments had a one day completion time limit. 

\paragraph{Platform}  Test platforms were: (a) CPU on a commodity laptop (2017 MacBook Pro); (b) CPU on a server (Xeon E5-2660, 40 hyperthreaded cores; 256GB RAM) with consumer GPU (NVIDIA GeForce GTX 1080 Ti); (c) datacentre GPU (NVIDIA Tesla P100 16GB). Platforms (b) and (c) both ran CUDA 10.1. SIMT CPU extensions such as AVX and AVX512 were left enabled or disabled according to {\tt \small pip} distribution defaults. 

\paragraph{Workload} Inference workloads are a pre-generated amount of random data. Scales are chosen to resemble the scales of real prediction workloads, especially for on-demand services. Inference set workload sizes range from $10^0$ to $10^8$ instances.

Code is available at {\tt \small https://github.com/leondz/inferencepower} (including full graphs).

\begin{figure}
    \centering
     \begin{subfigure}[b]{0.48\textwidth}
         \centering
         \includegraphics[width=\textwidth]{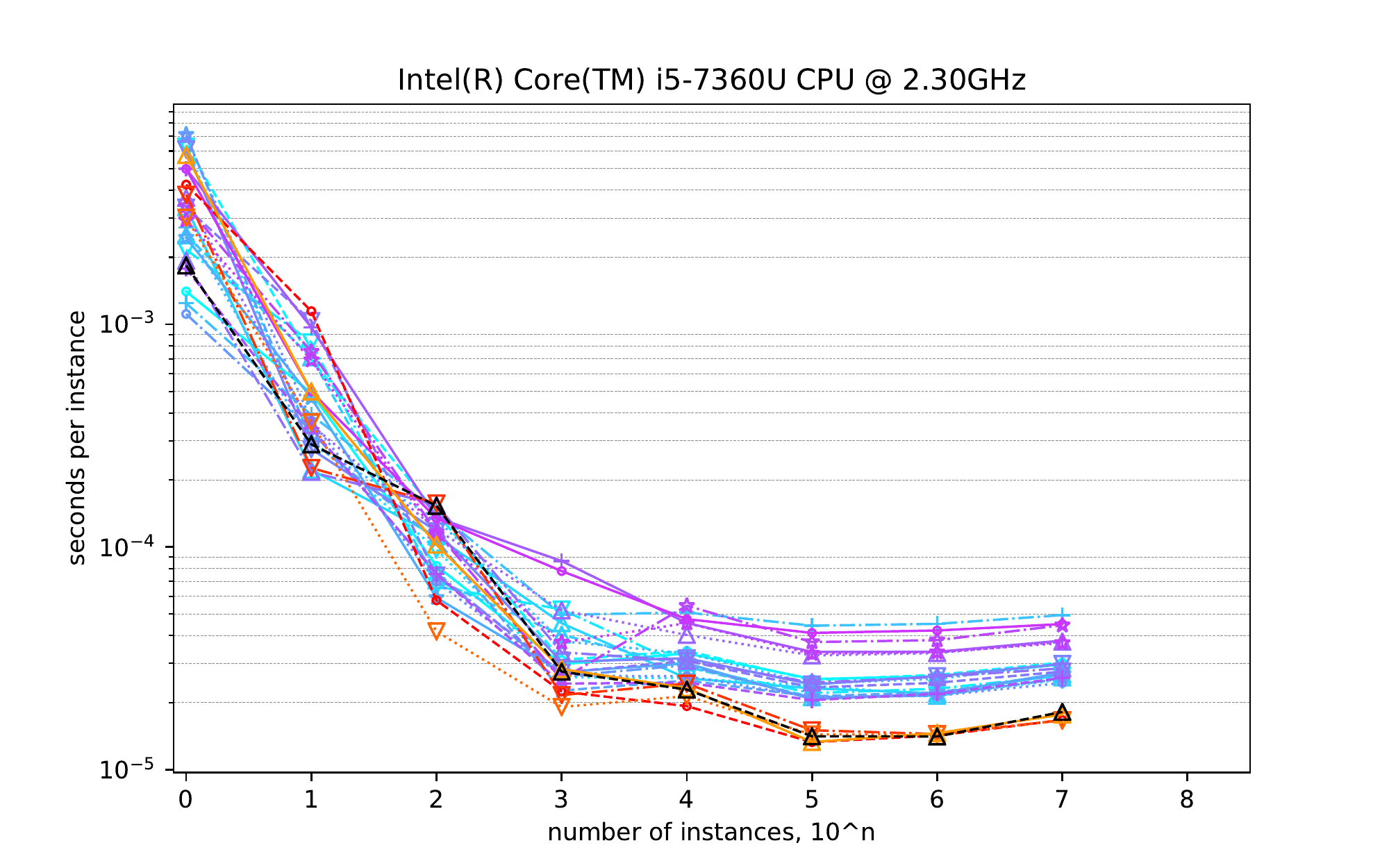}
         \caption{Consumer CPU}
         \label{fig:consumer_cpu}
     \end{subfigure}
     \hfill
     \begin{subfigure}[b]{0.48\textwidth}
         \centering
         \includegraphics[width=\textwidth]{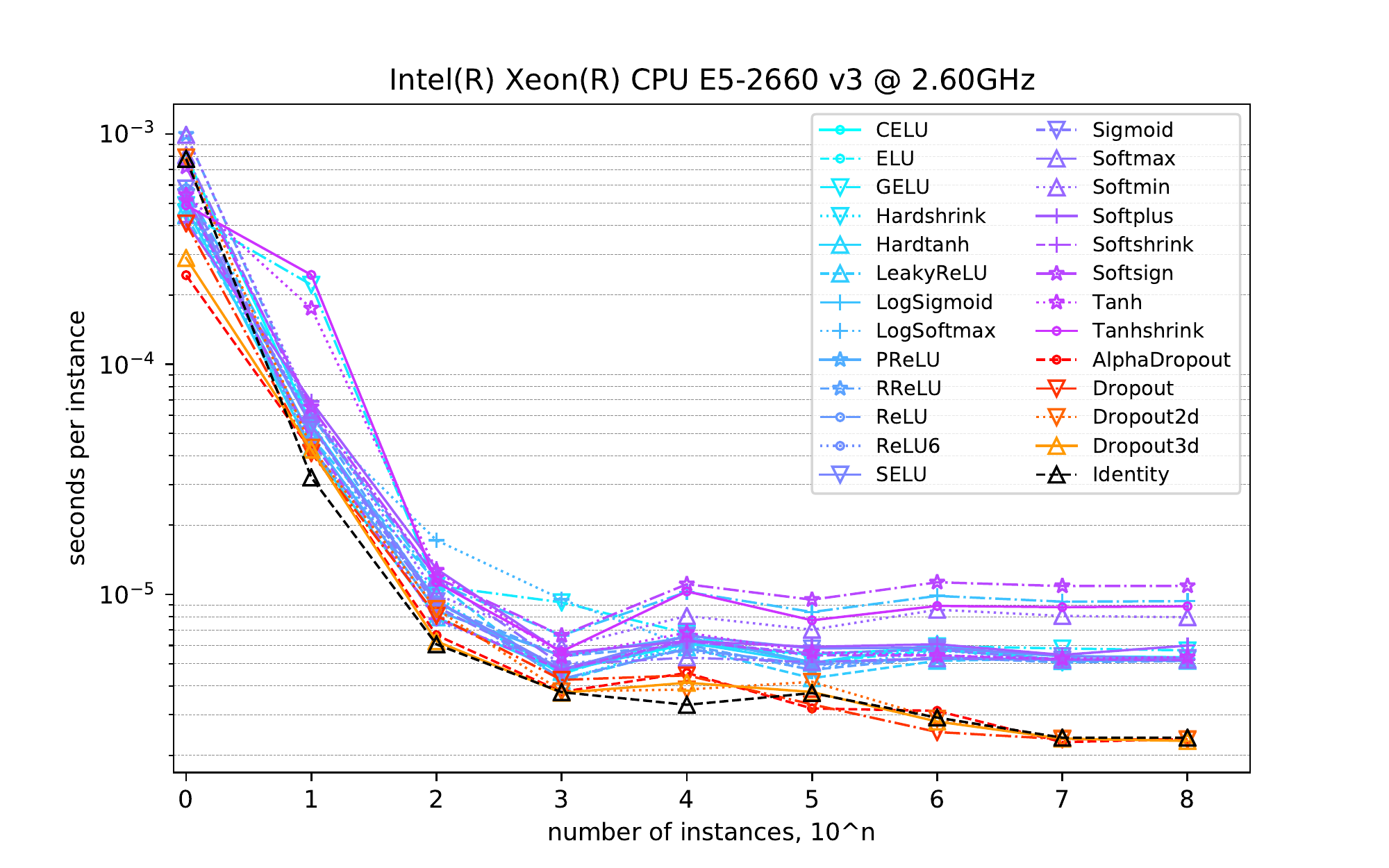}
         \caption{Datacentre CPU}
         \label{fig:dc_cpu}
     \end{subfigure}
     \hfill
     \begin{subfigure}[b]{0.48\textwidth}
         \centering
         \includegraphics[width=\textwidth]{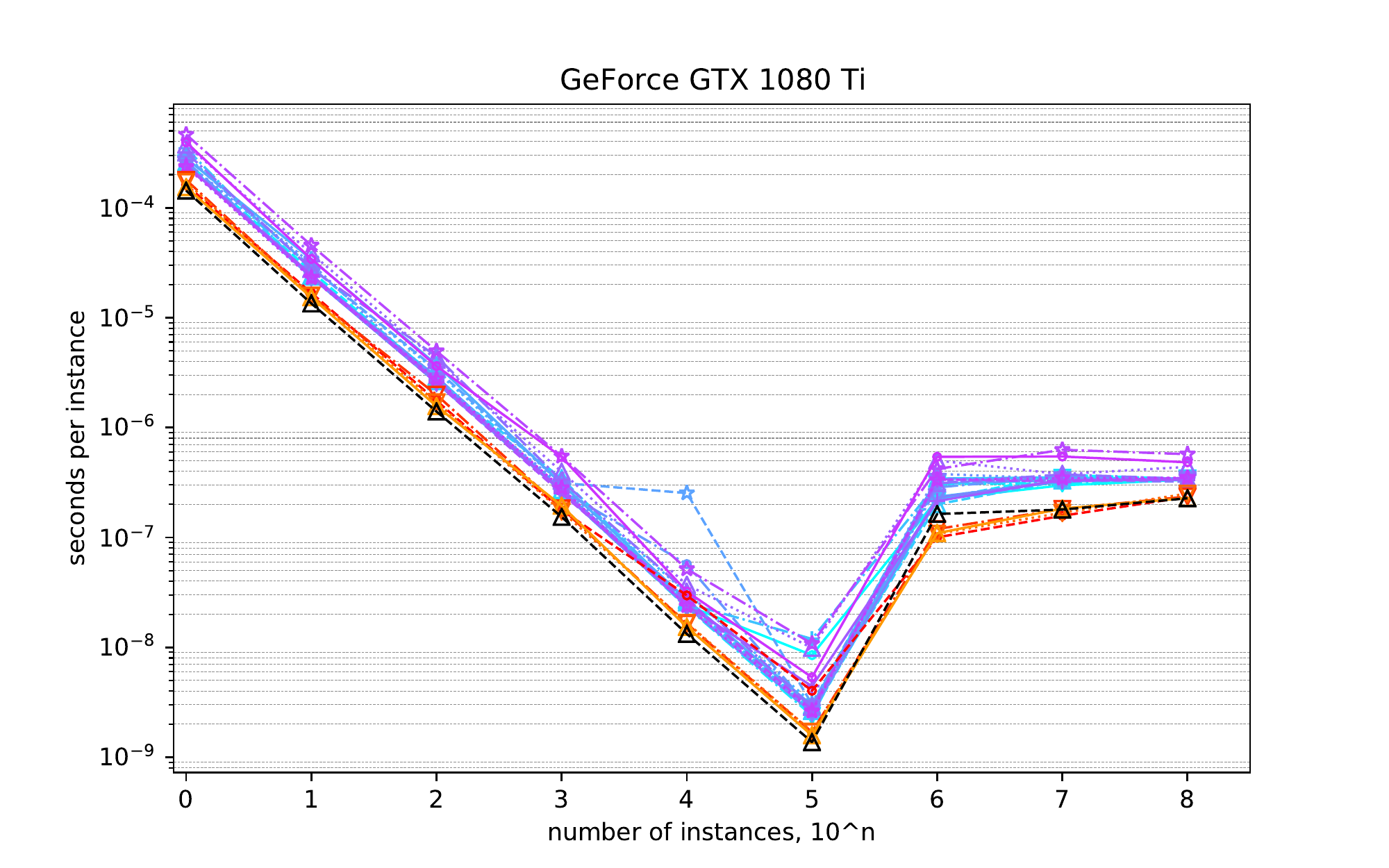}
         \caption{Consumer GPU}
         \label{fig:consumer_gpu}
     \end{subfigure}
     \hfill
     \begin{subfigure}[b]{0.48\textwidth}
         \centering
         \includegraphics[width=\textwidth]{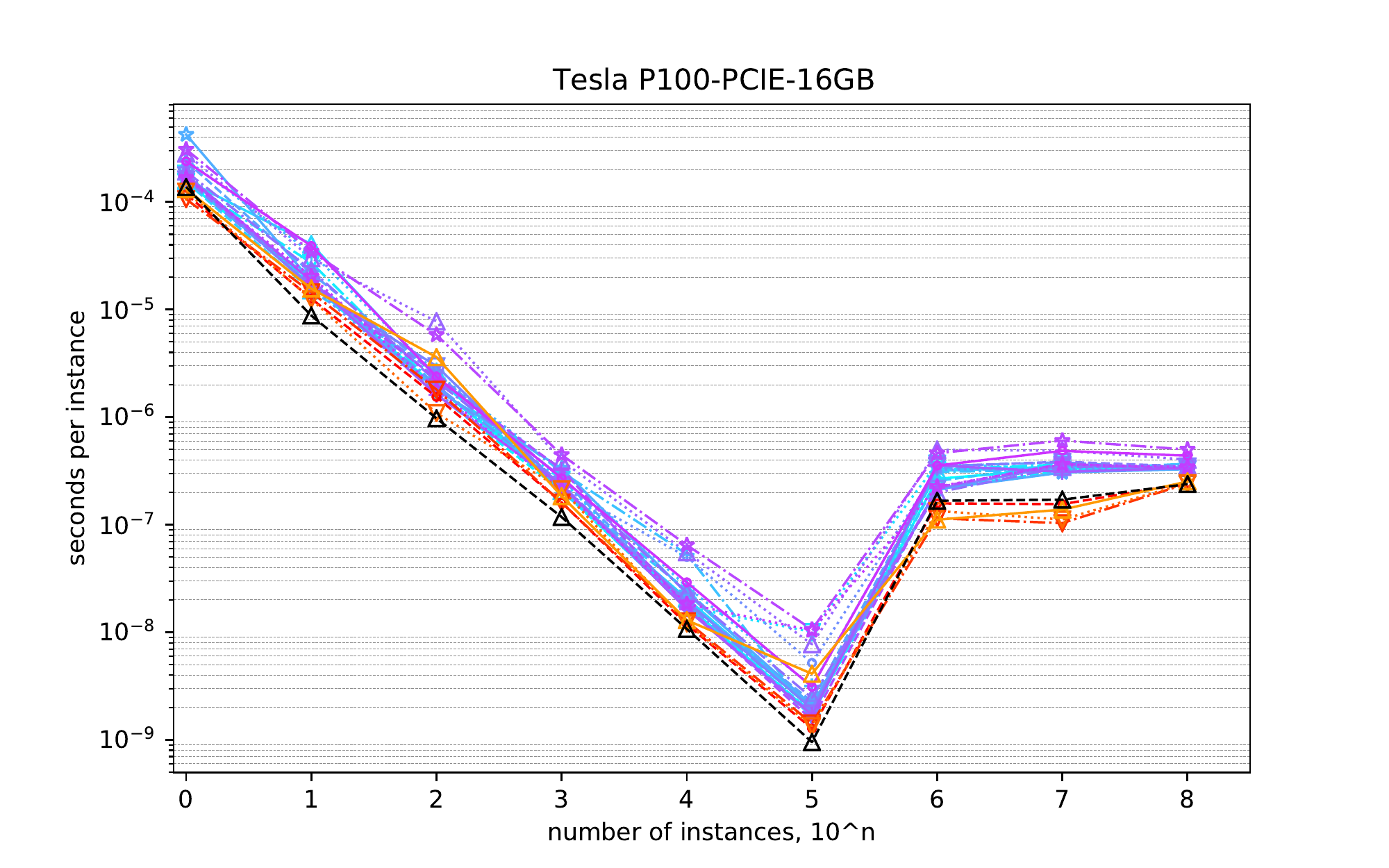}
         \caption{Datacentre GPU}
         \label{fig:dc_gpu}
     \end{subfigure}
    \caption{Per-instance function run time against inference data volume. Activation functions in blue-green colours, dropout in autumn colours, identity in grey.}
    \label{fig:timevsload}
\end{figure}

\section{Results and Discussion}
Figure~\ref{fig:timevsload} shows the inference time taken per-instance for a range of activation functions and inference set sizes. There are differences in the time taken by activation functions, and therefore power consumed. Function performance indicates that dropout functions have a lower power consumption, and that the identity function the lowest. The net time per instance is higher for smaller inference sets, which can be explained by the impact of fixed costs. 

Of the activation functions, tanh, logSigmoid, tanhshrink, and softmin are the slowest to run at scale. This indicates that, due to their increased estimated emissions impact,  use of these functions should be considered carefully before models using them are deployed broadly or deployed in an application with a long lifetime.

As the size of the inference set increases, so does the proportion of runtime spent running these functions. While the ``v''-shaped part of the curve suggests some caching/batching effects for the GPUs, the \textit{relative} difference between activation functions is the phenomenon of interest, and that persists. The spread between the fastest and slowest activation functions is roughly a factor of two, and is present across workload sizes and platform. There is a slight suggestion that the function-based efficiency spread may close gradually on GPUs with even larger inference sets. 

\begin{figure}[b]
    \centering
     \begin{subfigure}[t]{0.48\textwidth}
         \centering
         \includegraphics[width=\textwidth]{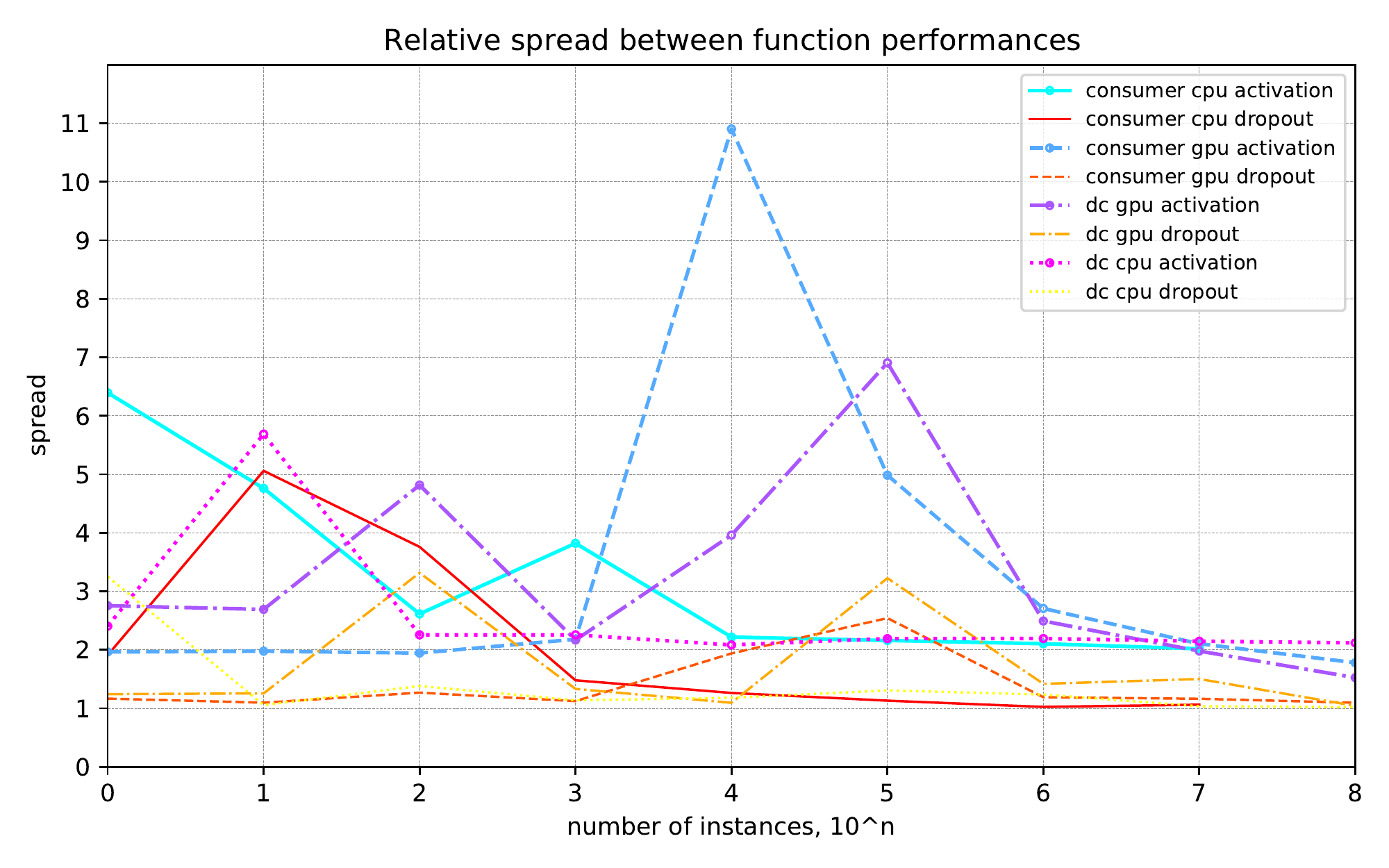}
         \caption{Intra-group spread}
         \label{fig:groupspread}
     \end{subfigure}
     \hfill
     \begin{subfigure}[t]{0.48\textwidth}
         \centering
         \includegraphics[width=\textwidth]{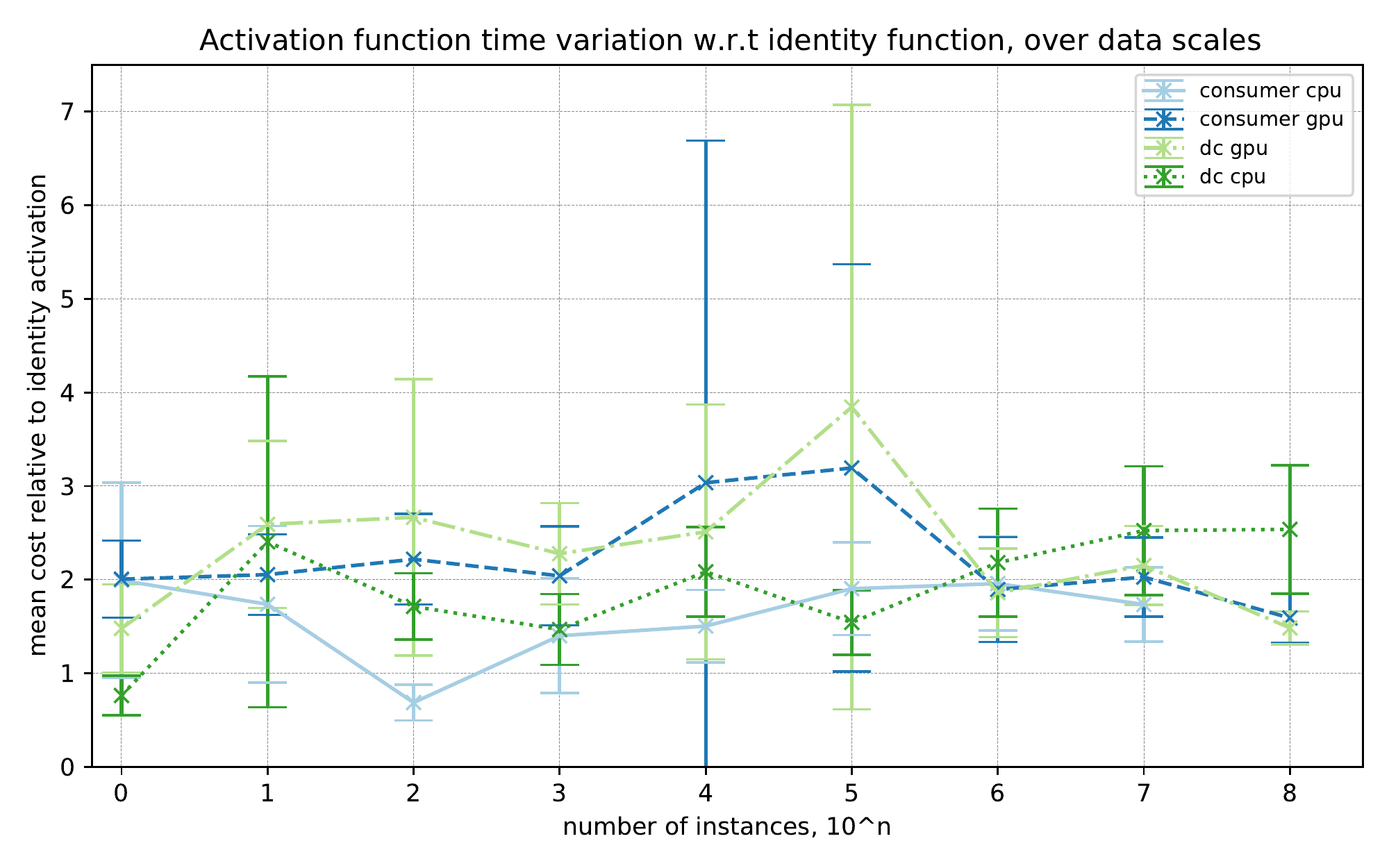}
         \caption{Activation function performance spread mean and s.d., relative to identity function (dropout functions excluded)}
         \label{fig:spreadidentity}
     \end{subfigure}

    \caption{Spread between high- and low-running time functions, over dataset scale and platforms.}
    \label{fig:spread}
\end{figure}

\paragraph{Performance varies between functions.} The scale of difference between fast and slow activation functions is shown  in Figure~\ref{fig:spread}. Dropout functions perform roughly as well as each other. The performance difference between activation functions varies, but the spread persists. CPU activation workload spreads are fairly consistent regardless of the size of the inference set. GPU spread varies depending on inference set size, with some smaller instances workload sizes presenting high variance, and a generally decreasing spread as inference set sizes rise. This suggests that, for GPUs, activation function choice has more effect in situations where inference is performed over smaller sets at a time. 

\paragraph{Inference workload size is important.} There are spikes in GPU spread at certain inference set scales. For example, the difference between activation functions on consumer GPU hardware was a factor of 11 when doing inference on a set of $10^4$ values, i.e. an order of magnitude. The datacentre GPU platform did inference on the workload of $10^5$ values seven times slower on the slowest activation function than on the fastest. The magnitude of the scale of variation indicates: (a) that applications should be analysed and tuned on the target hardware if one is to avoid particularly costly activation functions; (b) applications with high-frequency workloads of smaller inference sets may be particularly prone to raised power consumption and emissions due to activation function choice.

\paragraph{Consumption spreads are consistently present.} Figure~\ref{fig:spreadidentity} shows the mean and standard deviation in activation function performance across inference set sizes and platforms, normalised relative to the identity function's performance. Higher spreads and variations indicate greater potential impact from activation function tuning. GPUs processing time over different activation functions varies more than for CPUs, depending on inference set size. The size of the spread in absolute GPU timings seen at $10^4$ and $10^5$ (Figure~\ref{fig:groupspread} is echoed here. On the other hand, the consumer CPU platform experiences relatively little variation across activation function performances as inference set size increases. This suggests that tuning function choice in larger-scale machine learning environments, e.g. datacentres and GPU hardware, can lead to the greatest relative emission reductions.

\subsection{Training Effectiveness}

Low power activations are less useful if one needs to use them more often to do the same thing. This is especially important when training machine learning models. The number of iterations required is predicated upon not only network architecture, but also the training data, the hyperparameters, and the starting conditions. Further, depending on a model's usage scenario, the part of its total power consumption represented by training can be between everything (if one never predicts) and asymptotic to zero (if one does many predictions). Nevertheless, it is helpful to estimate the demands that different activations place during this phase.

To work out how many iterations are needed, a dummy workload and performance target can be set up. In this case, we used MNIST data~\citep{lecun1998gradient}. Th evaluation network was similar to that in Section~\ref{sec:setup}, but instead using the MNIST training data with an input dimension of 784, a final sigmoid output layer with ten nodes (one per digit), and optimised with stochastic gradient descent. The activation function of the middle four layers of 1024 nodes each is varied as the experiment parameter. The hardware is a server CPU, platform (b) from Section~\ref{sec:setup}. Training was stopped after the epoch when validation accuracy exceeded 0.90, or after 100 epochs. Time was only accumulated during training and not evaluation. Note that networks with hidden layers composed of regularising functions (i.e. dropout) and the identity function are still able to learn the target function in this setup due to the presence of sigmoid output function.

\begin{figure}[h!]
    \centering
    \includegraphics[width=\textwidth]{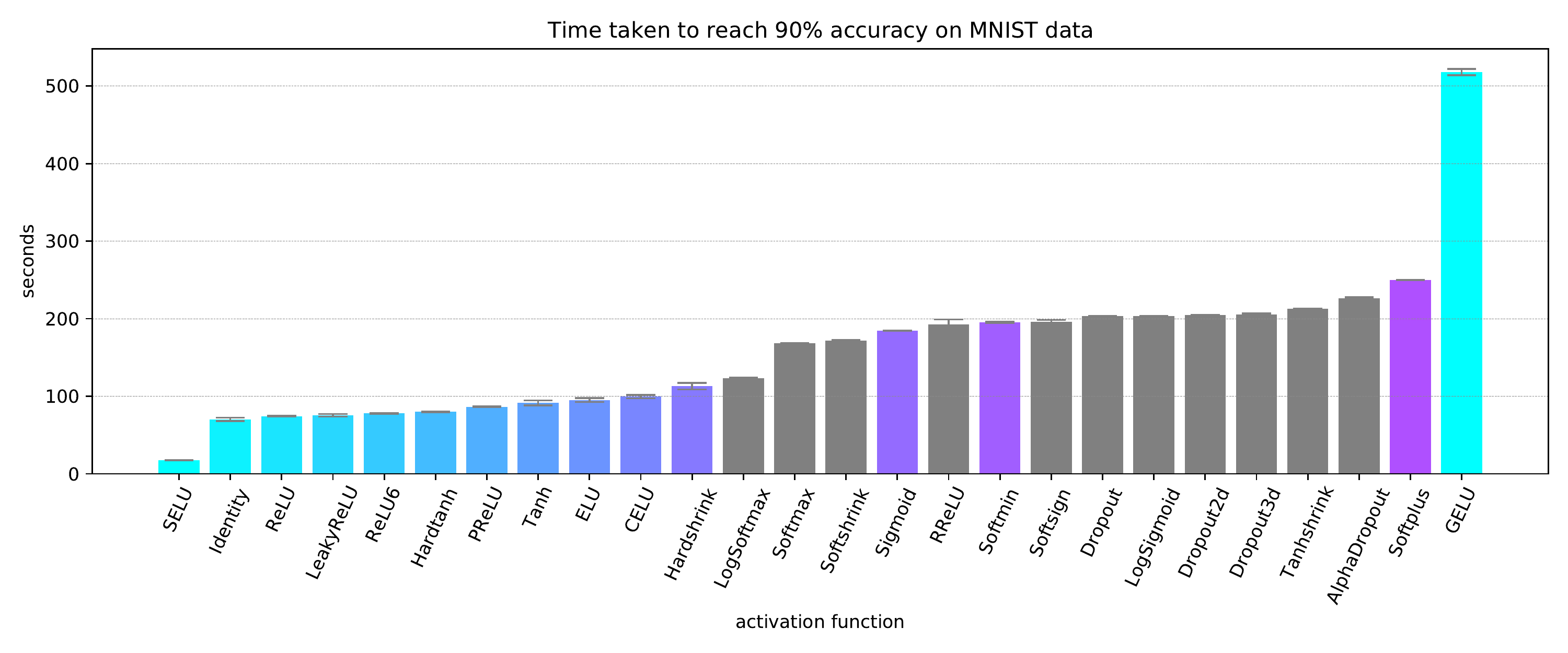}
    \caption{Time for functions to train on MNIST. Functions whose models do not reach the the required accuracy within 100 epochs have gray bars. Error bars are time standard deviation.}
    \label{fig:mnist}
\end{figure}

Figure~\ref{fig:mnist} shows the total time consumed to reach the target validation accuracy. Not all functions reached the required accuracy within the given number of epochs. If a function reached the maximum epoch count in any of its runs without achieving 90\% on the validation set, its bar is marked in grey. Scaled Exponential Linear Units~\citep{klambauer2017self} performed particularly well on this problem over multiple random initialisations. From a learning perspective, the networks have not performed particularly well: identity performed quickest, suggesting it was simpler to have a narrow sigmoid layer learn the problem than multiple broad hidden ones. Of the functions, linear units generally performed well; only one of these did not complete the problem in the required number of epochs. It is also possible that the chosen optimiser is not equally suitable for all activation functions. However, there is an indication that many of the functions that are efficient in earlier experiments evaluating inference-time performance can also perform well during training.

\section{Conclusion}
This paper estimated the power consumption of neural network activation functions. The range over activation functions was often a factor of two or more, with larger spreads for different platform-dependent workloads. This result was consistent across device type (CPU and GPU), on both consumer and datacenter hardware, and for various scales of dataset. The scale of spread indicates that choice of neural network activation function affects machine learning model power consumption.

\subsection*{Acknowledgments}
Thanks to Peter Sestoft for suggestions regarding hardware constraints. 

\bibliography{iclr2020_conference}
\bibliographystyle{iclr2020_conference}

\clearpage
\appendix
\section{Appendix}
\subsection{Timings for GPUs}

These are the mean absolute prediction times for various activation functions in seconds, on CUDA GPUs.

\begin{table}[h]
    \scriptsize
    \centering
    \begin{tabular}{lccccccccc}
    {\bf Function} & {\bf n=0} & {\bf n=1} & {\bf n=2} & {\bf n=3} & {\bf n=4} & {\bf n=5} & {\bf n=6} & {\bf n=7} & {\bf n=8} \\
\hline
CELU & 2.522e-04 & 2.389e-04 & 2.697e-04 & 2.696e-04 & 2.411e-04 & 8.489e-04 & 2.221e-01 & \textbf{2.981e+00} & 3.349e+01 \\
ELU & 2.626e-04 & 2.643e-04 & 2.794e-04 & 2.758e-04 & 2.460e-04 & 2.495e-04 & 2.213e-01 & 3.708e+00 & 3.400e+01 \\
GELU & 2.486e-04 & 2.346e-04 & 2.564e-04 & 2.617e-04 & \textbf{2.322e-04} & \textbf{2.391e-04} & 3.061e-01 & 3.110e+00 & 3.360e+01 \\
Hardshrink & 2.525e-04 & 2.418e-04 & 2.640e-04 & 2.739e-04 & 2.504e-04 & 2.846e-04 & 2.162e-01 & 3.490e+00 & 3.449e+01 \\
Hardtanh & 2.464e-04 & 2.401e-04 & 2.889e-04 & 2.690e-04 & 2.701e-04 & 2.954e-04 & 3.430e-01 & 3.522e+00 & 3.346e+01 \\
LeakyReLU & 2.570e-04 & 2.363e-04 & 2.633e-04 & 3.440e-04 & 2.557e-04 & 2.636e-04 & \textbf{1.999e-01} & 3.291e+00 & 3.352e+01 \\
LogSigmoid & 3.479e-04 & 2.349e-04 & 2.803e-04 & 2.611e-04 & 2.519e-04 & 1.192e-03 & 3.085e-01 & 3.303e+00 & 3.455e+01 \\
LogSoftmax & 2.970e-04 & 2.879e-04 & 3.233e-04 & 3.091e-04 & 3.112e-04 & 3.135e-04 & 2.302e-01 & 3.401e+00 & 3.369e+01 \\
PReLU & 2.814e-04 & 3.402e-04 & 3.737e-04 & 3.069e-04 & 2.715e-04 & 2.951e-04 & 2.341e-01 & 3.217e+00 & 3.420e+01 \\
RReLU & 2.920e-04 & 2.853e-04 & 3.370e-04 & 3.126e-04 & 2.532e-03 & 3.035e-04 & 3.447e-01 & 3.350e+00 & 3.354e+01 \\
ReLU & 2.656e-04 & 2.424e-04 & 2.884e-04 & 2.692e-04 & 5.717e-04 & 2.654e-04 & 2.855e-01 & 3.585e+00 & 3.333e+01 \\
ReLU6 & 2.590e-04 & 3.109e-04 & 2.555e-04 & 2.986e-04 & 2.356e-04 & 2.506e-04 & 3.785e-01 & 3.422e+00 & 3.409e+01 \\
SELU & 2.460e-04 & \textbf{2.335e-04} & 2.554e-04 & 2.617e-04 & 2.360e-04 & 2.543e-04 & 2.298e-01 & 3.304e+00 & 3.357e+01 \\
Sigmoid & 2.478e-04 & 2.457e-04 & 2.628e-04 & 2.725e-04 & 2.576e-04 & 2.758e-04 & 3.208e-01 & 3.307e+00 & \textbf{3.215e+01} \\
Softmax & 3.186e-04 & 2.774e-04 & 4.445e-04 & 3.273e-04 & 3.189e-04 & 2.874e-04 & 3.098e-01 & 3.742e+00 & 3.409e+01 \\
Softmin & 3.745e-04 & 3.840e-04 & 4.121e-04 & 3.955e-04 & 3.710e-04 & 9.869e-04 & 5.005e-01 & 3.764e+00 & 4.382e+01 \\
Softplus & 2.444e-04 & 2.448e-04 & 2.671e-04 & 2.652e-04 & 2.412e-04 & 4.425e-04 & 2.144e-01 & 3.332e+00 & 3.343e+01 \\
Softshrink & 2.396e-04 & 2.373e-04 & \textbf{2.554e-04} & \textbf{2.510e-04} & 2.736e-04 & 2.700e-04 & 3.378e-01 & 3.232e+00 & 3.488e+01 \\
Softsign & 4.635e-04 & 4.544e-04 & 4.967e-04 & 5.473e-04 & 5.158e-04 & 1.079e-03 & 4.194e-01 & 6.273e+00 & 5.721e+01 \\
Tanh & \textbf{2.360e-04} & 2.298e-04 & 2.766e-04 & 2.810e-04 & 2.385e-04 & 2.621e-04 & 3.345e-01 & 3.458e+00 & 3.495e+01 \\
Tanhshrink & 3.942e-04 & 3.417e-04 & 3.621e-04 & 5.387e-04 & 3.222e-04 & 5.335e-04 & 5.413e-01 & 5.447e+00 & 4.826e+01 \\
\hline
AlphaDropout & 1.613e-04 & 1.677e-04 & 1.754e-04 & 1.774e-04 & 2.967e-04 & 4.021e-04 & \textbf{1.002e-01} & \textbf{1.574e+00} & 2.337e+01 \\
Dropout & 1.791e-04 & 1.584e-04 & 1.988e-04 & 1.833e-04 & \textbf{1.641e-04} & 1.693e-04 & 1.192e-01 & 1.815e+00 & 2.353e+01 \\
Dropout2d & 1.713e-04 & 1.585e-04 & \textbf{1.695e-04} & \textbf{1.728e-04} & 1.672e-04 & 1.706e-04 & 1.079e-01 & 1.675e+00 & 2.500e+01 \\
Dropout3d & \textbf{1.535e-04} & \textbf{1.529e-04} & 1.567e-04 & 1.940e-04 & 1.531e-04 & \textbf{1.582e-04} & 1.093e-01 & 1.831e+00 & \textbf{2.283e+01} \\
\hline
Identity & 1.432e-04 & 1.346e-04 & 1.398e-04 & 1.538e-04 & 1.325e-04 & 1.371e-04 & 1.634e-01 & 1.791e+00 & 2.271e+01 \\
\hline
    \end{tabular}
    \caption{For GTX-1080i, $10^n$ instances}
    \label{tab:consumer_gpu}
\end{table}

\begin{table}[h]
    \scriptsize
    \centering
    \begin{tabular}{lccccccccc}
    {\bf Function} & {\bf n=0} & {\bf n=1} & {\bf n=2} & {\bf n=3} & {\bf n=4} & {\bf n=5} & {\bf n=6} & {\bf n=7} & {\bf n=8} \\
\hline
CELU & 1.543e-04 & 1.835e-04 & 1.909e-04 & 2.285e-04 & 1.789e-04 & 1.758e-04 & 2.672e-01 & 3.320e+00 & 3.397e+01 \\
ELU & 1.815e-04 & 1.756e-04 & 1.637e-04 & 2.072e-04 & 1.733e-04 & 1.797e-04 & 2.155e-01 & 3.171e+00 & 3.289e+01 \\
GELU & 1.538e-04 & 2.726e-04 & 1.931e-04 & 2.135e-04 & 1.700e-04 & 1.666e-04 & 3.252e-01 & 3.635e+00 & 3.303e+01 \\
Hardshrink & 1.842e-04 & 1.683e-04 & 1.848e-04 & 2.626e-04 & 1.724e-04 & 1.008e-03 & 3.703e-01 & 3.278e+00 & 3.311e+01 \\
Hardtanh & 1.595e-04 & 4.051e-04 & 2.345e-04 & 2.061e-04 & 2.063e-04 & 2.024e-04 & 3.164e-01 & 3.479e+00 & 3.363e+01 \\
LeakyReLU & \textbf{1.539e-04} & \textbf{1.505e-04} & 2.635e-04 & \textbf{2.060e-04} & 2.474e-04 & 1.916e-04 & 2.553e-01 & 3.624e+00 & 3.378e+01 \\
LogSigmoid & 1.635e-04 & 1.654e-04 & 2.347e-04 & 3.102e-04 & 5.318e-04 & 1.731e-04 & 3.101e-01 & 3.234e+00 & 3.693e+01 \\
LogSoftmax & 2.482e-04 & 3.350e-04 & 2.560e-04 & 3.167e-04 & 2.648e-04 & 3.161e-04 & 3.525e-01 & 3.134e+00 & 3.424e+01 \\
PReLU & 4.236e-04 & 1.844e-04 & 1.808e-04 & 3.335e-04 & 2.307e-04 & 2.096e-04 & 2.138e-01 & \textbf{3.060e+00} & 3.284e+01 \\
RReLU & 2.388e-04 & 2.179e-04 & 2.544e-04 & 2.689e-04 & 1.894e-04 & 2.273e-04 & 3.394e-01 & 3.127e+00 & 3.404e+01 \\
ReLU & 1.643e-04 & 1.906e-04 & 1.971e-04 & 2.359e-04 & 1.841e-04 & 1.953e-04 & 3.507e-01 & 3.804e+00 & \textbf{3.280e+01} \\
ReLU6 & 1.858e-04 & 1.632e-04 & 1.709e-04 & 2.246e-04 & 5.023e-04 & 5.198e-04 & 3.258e-01 & 3.304e+00 & 3.410e+01 \\
SELU & 1.775e-04 & 1.608e-04 & 2.979e-04 & 2.239e-04 & 1.946e-04 & 1.946e-04 & 2.156e-01 & 3.562e+00 & 3.309e+01 \\
Sigmoid & 1.618e-04 & 1.677e-04 & 2.544e-04 & 2.363e-04 & 1.690e-04 & 1.746e-04 & 3.489e-01 & 3.851e+00 & 3.487e+01 \\
Softmax & 1.917e-04 & 2.327e-04 & 2.146e-04 & 2.772e-04 & 2.345e-04 & 2.459e-04 & \textbf{1.993e-01} & 3.438e+00 & 3.517e+01 \\
Softmin & 2.812e-04 & 3.000e-04 & 7.762e-04 & 3.937e-04 & 5.543e-04 & 7.737e-04 & 4.968e-01 & 4.876e+00 & 4.034e+01 \\
Softplus & 1.779e-04 & 1.845e-04 & \textbf{1.612e-04} & 2.331e-04 & \textbf{1.634e-04} & 1.702e-04 & 3.565e-01 & 3.095e+00 & 3.366e+01 \\
Softshrink & 1.719e-04 & 1.843e-04 & 2.281e-04 & 3.273e-04 & 1.721e-04 & \textbf{1.537e-04} & 2.228e-01 & 3.688e+00 & 3.308e+01 \\
Softsign & 3.077e-04 & 3.466e-04 & 5.812e-04 & 4.464e-04 & 6.471e-04 & 1.061e-03 & 4.625e-01 & 6.055e+00 & 5.002e+01 \\
Tanh & 1.671e-04 & 2.035e-04 & 1.817e-04 & 2.137e-04 & 1.838e-04 & 1.041e-03 & 2.259e-01 & 3.637e+00 & 3.433e+01 \\
Tanhshrink & 2.387e-04 & 3.910e-04 & 2.380e-04 & 2.706e-04 & 2.918e-04 & 3.118e-04 & 3.549e-01 & 4.865e+00 & 4.383e+01 \\
\hline
AlphaDropout & 1.179e-04 & 1.262e-04 & 1.541e-04 & 1.613e-04 & \textbf{1.184e-04} & \textbf{1.276e-04} & 1.579e-01 & 1.553e+00 & 2.434e+01 \\
Dropout & \textbf{1.062e-04} & 1.445e-04 & 1.803e-04 & \textbf{1.607e-04} & 1.267e-04 & 1.424e-04 & 1.147e-01 & \textbf{1.035e+00} & \textbf{2.419e+01} \\
Dropout2d & 1.251e-04 & \textbf{1.247e-04} & \textbf{1.090e-04} & 2.141e-04 & 1.218e-04 & 1.338e-04 & 1.341e-01 & 1.119e+00 & 2.441e+01 \\
Dropout3d & 1.319e-04 & 1.565e-04 & 3.613e-04 & 1.842e-04 & 1.296e-04 & 4.115e-04 & \textbf{1.113e-01} & 1.382e+00 & 2.522e+01 \\
\hline
Identity & 1.380e-04 & 8.792e-05 & 9.750e-05 & 1.179e-04 & 1.074e-04 & 9.531e-05 & 1.671e-01 & 1.710e+00 & 2.386e+01 \\
\hline

    \end{tabular}
    \caption{For Tesla P100, $10^n$ instances}
    \label{tab:dc_gpu}
\end{table}

\clearpage
\subsection{Timings for CPUs}

These are the mean absolute prediction times for various activation functions in seconds, on CPUs.

\begin{table}[h]
\scriptsize
    \centering
    \begin{tabular}{lccccccccc}
    {\bf Function} & {\bf n=0} & {\bf n=1} & {\bf n=2} & {\bf n=3} & {\bf n=4} & {\bf n=5} & {\bf n=6} & {\bf n=7} & {\bf n=8} \\
\hline
CELU & 1.405e-03 & 4.911e-03 & 8.225e-03 & 2.975e-02 & 3.330e-01 & 2.553e+00 & 2.626e+01 & 2.969e+02 & 0.000e+00 \\
ELU & 6.384e-03 & 6.949e-03 & 1.421e-02 & 3.112e-02 & 3.414e-01 & 2.527e+00 & 2.664e+01 & 3.030e+02 & 0.000e+00 \\
GELU & 2.163e-03 & 8.301e-03 & 6.664e-03 & 5.240e-02 & 2.892e-01 & 2.206e+00 & 2.309e+01 & 2.645e+02 & 0.000e+00 \\
Hardshrink & 6.289e-03 & 3.041e-03 & 9.742e-03 & 2.602e-02 & 2.575e-01 & 2.383e+00 & 2.158e+01 & 2.537e+02 & 0.000e+00 \\
Hardtanh & 3.353e-03 & 2.221e-03 & 1.117e-02 & 4.544e-02 & 2.571e-01 & 2.318e+00 & 2.202e+01 & 2.683e+02 & 0.000e+00 \\
LeakyReLU & 2.511e-03 & 7.106e-03 & 7.090e-03 & 4.036e-02 & 2.896e-01 & 2.116e+00 & 2.146e+01 & 2.593e+02 & 0.000e+00 \\
LogSigmoid & 1.243e-03 & 3.914e-03 & 1.446e-02 & 4.962e-02 & 5.080e-01 & 4.431e+00 & 4.517e+01 & 4.945e+02 & 0.000e+00 \\
LogSoftmax & 2.719e-03 & 2.893e-03 & 1.333e-02 & 3.773e-02 & 3.329e-01 & 2.436e+00 & 2.557e+01 & 2.985e+02 & 0.000e+00 \\
PReLU & 2.453e-03 & 4.674e-03 & \textbf{5.918e-03} & 2.744e-02 & 3.002e-01 & 2.072e+00 & 2.201e+01 & 2.721e+02 & 0.000e+00 \\
RReLU & 7.116e-03 & 2.999e-03 & 1.075e-02 & \textbf{2.264e-02} & 2.514e-01 & 2.135e+00 & 2.223e+01 & 2.685e+02 & 0.000e+00 \\
ReLU & \textbf{1.113e-03} & 3.330e-03 & 7.654e-03 & 2.626e-02 & 2.948e-01 & 2.087e+00 & 2.165e+01 & 2.564e+02 & 0.000e+00 \\
ReLU6 & 6.967e-03 & 3.698e-03 & 6.990e-03 & 2.638e-02 & 2.631e-01 & 2.117e+00 & \textbf{2.148e+01} & \textbf{2.451e+02} & 0.000e+00 \\
SELU & 6.106e-03 & 2.751e-03 & 1.178e-02 & 3.055e-02 & 3.165e-01 & 2.414e+00 & 2.629e+01 & 2.985e+02 & 0.000e+00 \\
Sigmoid & 3.347e-03 & 1.035e-02 & 7.452e-03 & 2.728e-02 & 3.080e-01 & 2.338e+00 & 2.456e+01 & 2.757e+02 & 0.000e+00 \\
Softmax & 1.927e-03 & \textbf{2.174e-03} & 1.547e-02 & 3.365e-02 & 3.083e-01 & 2.462e+00 & 2.610e+01 & 2.850e+02 & 0.000e+00 \\
Softmin & 3.691e-03 & 3.531e-03 & 1.200e-02 & 5.189e-02 & 4.026e-01 & 3.259e+00 & 3.337e+01 & 3.755e+02 & 0.000e+00 \\
Softplus & 4.992e-03 & 9.673e-03 & 1.355e-02 & 8.653e-02 & 4.558e-01 & 3.379e+00 & 3.382e+01 & 3.799e+02 & 0.000e+00 \\
Softshrink & 1.781e-03 & 3.413e-03 & 7.560e-03 & 2.436e-02 & \textbf{2.457e-01} & \textbf{2.052e+00} & 2.192e+01 & 2.582e+02 & 0.000e+00 \\
Softsign & 3.399e-03 & 7.535e-03 & 1.170e-02 & 2.544e-02 & 5.449e-01 & 3.742e+00 & 3.815e+01 & 4.451e+02 & 0.000e+00 \\
Tanh & 2.939e-03 & 6.905e-03 & 1.137e-02 & 3.721e-02 & 4.553e-01 & 3.319e+00 & 3.366e+01 & 3.707e+02 & 0.000e+00 \\
Tanhshrink & 4.985e-03 & 4.995e-03 & 1.345e-02 & 7.797e-02 & 4.744e-01 & 4.112e+00 & 4.218e+01 & 4.508e+02 & 0.000e+00 \\
\hline
AlphaDropout & 4.242e-03 & \textbf{1.146e-02} & 5.761e-03 & 2.243e-02 & \textbf{1.929e-01} & \textbf{1.329e+00} & \textbf{1.424e+01} & 1.672e+02 & 0.000e+00 \\
Dropout & 3.823e-03 & 2.265e-03 & 1.571e-02 & 2.163e-02 & 2.436e-01 & 1.503e+00 & 1.445e+01 & 1.667e+02 & 0.000e+00 \\
Dropout2d & \textbf{3.012e-03} & 3.630e-03 & \textbf{4.178e-03} & \textbf{1.913e-02} & 2.140e-01 & 1.439e+00 & 1.440e+01 & \textbf{1.665e+02} & 0.000e+00 \\
Dropout3d & 5.751e-03 & 4.988e-03 & 1.025e-02 & 2.829e-02 & 2.323e-01 & 1.331e+00 & 1.458e+01 & 1.768e+02 & 0.000e+00 \\
\hline
Identity & 1.838e-03 & 2.890e-03 & 1.530e-02 & 2.755e-02 & 2.290e-01 & 1.413e+00 & 1.409e+01 & 1.818e+02 & 0.000e+00 \\
\hline
    \end{tabular}
    \caption{For MacBook Pro 2017 / i5-7360U, $10^n$ instances ($n=8$ did not complete in time)}
    \label{tab:consumer_cpu}
\end{table}

\begin{table}[h]
    \scriptsize
    \centering
    \begin{tabular}{lccccccccc}
    {\bf Function} & {\bf n=0} & {\bf n=1} & {\bf n=2} & {\bf n=3} & {\bf n=4} & {\bf n=5} & {\bf n=6} & {\bf n=7} & {\bf n=8} \\
\hline
CELU & 8.040e-04 & 5.504e-04 & 9.299e-04 & 4.652e-03 & 6.080e-02 & 5.491e-01 & 5.843e+00 & 5.390e+01 & 5.333e+02 \\
ELU & 5.142e-04 & 5.420e-04 & 9.370e-04 & 4.503e-03 & 6.701e-02 & 5.052e-01 & 5.739e+00 & 5.195e+01 & 5.303e+02 \\
GELU & 4.563e-04 & 2.212e-03 & 1.090e-03 & 9.235e-03 & 6.776e-02 & 5.033e-01 & 5.956e+00 & 5.824e+01 & 5.713e+02 \\
Hardshrink & 4.909e-04 & 4.580e-04 & 9.830e-04 & 4.254e-03 & 6.083e-02 & 5.435e-01 & 5.480e+00 & 5.065e+01 & 5.231e+02 \\
Hardtanh & 4.804e-04 & \textbf{4.295e-04} & 7.947e-04 & 4.607e-03 & 6.192e-02 & 5.038e-01 & 6.082e+00 & 5.125e+01 & 5.213e+02 \\
LeakyReLU & 4.435e-04 & 4.814e-04 & 1.130e-03 & \textbf{4.219e-03} & 5.889e-02 & 4.327e-01 & \textbf{5.147e+00} & 5.265e+01 & 5.146e+02 \\
LogSigmoid & 5.823e-04 & 6.052e-04 & 1.124e-03 & 6.629e-03 & 1.026e-01 & 8.386e-01 & 9.849e+00 & 9.301e+01 & 9.348e+02 \\
LogSoftmax & 4.839e-04 & 5.797e-04 & 1.719e-03 & 9.518e-03 & 5.974e-02 & 4.708e-01 & 5.720e+00 & 5.131e+01 & 5.260e+02 \\
PReLU & 5.169e-04 & 4.822e-04 & 8.476e-04 & 5.408e-03 & 6.506e-02 & 5.080e-01 & 5.241e+00 & 5.224e+01 & 5.137e+02 \\
RReLU & 9.946e-04 & 6.080e-04 & 9.066e-04 & 5.379e-03 & 6.046e-02 & \textbf{4.700e-01} & 5.331e+00 & \textbf{5.063e+01} & \textbf{5.138e+02} \\
ReLU & 5.820e-04 & \textbf{4.295e-04} & 9.473e-04 & 4.295e-03 & 5.799e-02 & 4.943e-01 & 5.329e+00 & 5.182e+01 & 5.199e+02 \\
ReLU6 & 5.297e-04 & 5.645e-04 & \textbf{7.616e-04} & 4.971e-03 & 5.682e-02 & 4.900e-01 & 5.288e+00 & 5.230e+01 & 5.143e+02 \\
SELU & 5.307e-04 & 5.419e-04 & 9.343e-04 & 4.726e-03 & 6.620e-02 & 5.824e-01 & 5.878e+00 & 5.397e+01 & 5.294e+02 \\
Sigmoid & 5.802e-04 & 5.382e-04 & 8.845e-04 & 4.689e-03 & 6.233e-02 & 5.550e-01 & 5.756e+00 & 5.209e+01 & 5.282e+02 \\
Softmax & 8.080e-04 & 6.075e-04 & 1.183e-03 & 4.950e-03 & \textbf{5.313e-02} & 5.086e-01 & 5.270e+00 & 5.189e+01 & 5.189e+02 \\
Softmin & 9.904e-04 & 6.343e-04 & 1.002e-03 & 5.897e-03 & 8.061e-02 & 7.023e-01 & 8.580e+00 & 8.063e+01 & 7.957e+02 \\
Softplus & \textbf{4.135e-04} & 6.887e-04 & 1.290e-03 & 5.570e-03 & 6.268e-02 & 5.921e-01 & 6.075e+00 & 5.461e+01 & 5.993e+02 \\
Softshrink & 5.241e-04 & 4.692e-04 & 7.834e-04 & 4.706e-03 & 6.308e-02 & 5.442e-01 & 5.429e+00 & 5.219e+01 & 5.183e+02 \\
Softsign & 7.190e-04 & 6.517e-04 & 1.181e-03 & 6.631e-03 & 1.107e-01 & 9.482e-01 & 1.129e+01 & 1.087e+02 & 1.089e+03 \\
Tanh & 5.437e-04 & 1.747e-03 & 1.270e-03 & 5.356e-03 & 6.823e-02 & 5.610e-01 & 5.432e+00 & 5.245e+01 & 5.342e+02 \\
Tanhshrink & 4.899e-04 & 2.443e-03 & 1.129e-03 & 5.651e-03 & 1.030e-01 & 7.718e-01 & 8.912e+00 & 8.803e+01 & 8.879e+02 \\
\hline
AlphaDropout & \textbf{2.436e-04} & 4.216e-04 & 6.650e-04 & 3.773e-03 & 4.560e-02 & \textbf{3.195e-01} & 3.119e+00 & \textbf{2.281e+01} & 2.353e+02 \\
Dropout & 4.084e-04 & \textbf{4.113e-04} & 8.083e-04 & 4.251e-03 & 4.448e-02 & 3.329e-01 & \textbf{2.524e+00} & 2.351e+01 & 2.342e+02 \\
Dropout2d & 7.918e-04 & 4.341e-04 & 8.644e-04 & 3.793e-03 & \textbf{3.861e-02} & 4.170e-01 & 2.883e+00 & 2.364e+01 & 2.349e+02 \\
Dropout3d & 2.893e-04 & 4.250e-04 & \textbf{6.257e-04} & \textbf{3.737e-03} & 4.131e-02 & 3.767e-01 & 2.803e+00 & 2.363e+01 & \textbf{2.309e+02} \\
\hline
Identity & 7.799e-04 & 3.222e-04 & 6.066e-04 & 3.761e-03 & 3.314e-02 & 3.727e-01 & 2.917e+00 & 2.386e+01 & 2.387e+02 \\
\hline

    \end{tabular}
    \caption{For Intel Xeon E5-2660, $10^n$ instances}
    \label{tab:dc_cpu}
\end{table}

\end{document}

%% file: math_commands.tex
\usepackage{amsmath,amsfonts,bm}

\def\eqref#1{equation~\ref{#1}}

\def\1{\bm{1}}

\DeclareMathAlphabet{\mathsfit}{\encodingdefault}{\sfdefault}{m}{sl}
\SetMathAlphabet{\mathsfit}{bold}{\encodingdefault}{\sfdefault}{bx}{n}













%% file: activation_power.bbl
\begin{thebibliography}{19}
\providecommand{\natexlab}[1]{#1}
\providecommand{\url}[1]{\texttt{#1}}
\expandafter\ifx\csname urlstyle\endcsname\relax
  \providecommand{\doi}[1]{doi: #1}\else
  \providecommand{\doi}{doi: \begingroup \urlstyle{rm}\Url}\fi

\bibitem[Alvarez \& Park(2019)Alvarez and Park]{alvarez2019end}
Raziel Alvarez and Hyun-Jin Park.
\newblock End-to-end streaming keyword spotting.
\newblock In \emph{Proceedings of the International Conference on Acoustics,
  Speech and Signal Processing (ICASSP)}, pp.\  6336--6340. IEEE, 2019.

\bibitem[Amodei \& Hernandez(2018)Amodei and Hernandez]{amodei18}
Dario Amodei and Danny Hernandez.
\newblock {AI and C}ompute.
\newblock https://openai.com/blog/ai-and-compute/, 2018.
\newblock OpenAI.

\bibitem[Durant et~al.(2017)Durant, Giroux, Harris, and Stam]{durant2017inside}
Luke Durant, Olivier Giroux, Mark Harris, and Nick Stam.
\newblock Inside {Volta: T}he world’s most advanced data center {GPU}.
\newblock \emph{NVidia Parallel for All Blog,
  https://devblogs.nvidia.com/parallelforall/inside-volta}, 2017.

\bibitem[Fan et~al.(2020)Fan, Glavaš, Joty, Moosav, Shwartz, Wang, and
  Wolf]{sustainlp2020}
Angela Fan, Goran Glavaš, Shafiq Joty, Nafise~Sadat Moosav, Vered Shwartz,
  Alex Wang, and Thomas Wolf.
\newblock Proceedings of the first workshop on simple and efficient natural
  language processing.
\newblock In \emph{Workshops of the Conference on Empirical Methods in Natural
  Language Processing (EMNLP)}. ACL, 2020.

\bibitem[Fog(2019)]{agner}
Agner Fog.
\newblock Instruction tables.
\newblock https://www.agner.org/optimize/, 2019.
\newblock Danmarks Tekniske Universitet.

\bibitem[Girosi et~al.(1995)Girosi, Jones, and
  Poggio]{girosi1995regularization}
Federico Girosi, Michael Jones, and Tomaso Poggio.
\newblock Regularization theory and neural networks architectures.
\newblock \emph{Neural Computation}, 7\penalty0 (2):\penalty0 219--269, 1995.

\bibitem[Henderson et~al.(2020)Henderson, Hu, Romoff, Brunskill, Jurafsky, and
  Pineau]{henderson2020towards}
Peter Henderson, Jieru Hu, Joshua Romoff, Emma Brunskill, Dan Jurafsky, and
  Joelle Pineau.
\newblock Towards the systematic reporting of the energy and carbon footprints
  of machine learning.
\newblock \emph{arXiv}, abs/2002.05651, 2020.

\bibitem[Kim \& Rush(2016)Kim and Rush]{kim2016sequence}
Yoon Kim and Alexander~M Rush.
\newblock Sequence-level knowledge distillation.
\newblock In \emph{Proceedings of the Conference on Empirical Methods in
  Natural Language Processing (EMNLP)}, pp.\  1317--1327. ACL, 2016.

\bibitem[Klambauer et~al.(2017)Klambauer, Unterthiner, Mayr, and
  Hochreiter]{klambauer2017self}
G{\"u}nter Klambauer, Thomas Unterthiner, Andreas Mayr, and Sepp Hochreiter.
\newblock Self-normalizing neural networks.
\newblock In \emph{Advances in neural information processing systems}, pp.\
  971--980, 2017.

\bibitem[Krizhevsky et~al.(2012)Krizhevsky, Sutskever, and
  Hinton]{krizhevsky2012imagenet}
Alex Krizhevsky, Ilya Sutskever, and Geoffrey~E Hinton.
\newblock Imagenet classification with deep convolutional neural networks.
\newblock In \emph{Advances in neural information processing systems}, pp.\
  1097--1105, 2012.

\bibitem[LeCun et~al.(1998)LeCun, Bottou, Bengio, and
  Haffner]{lecun1998gradient}
Yann LeCun, L{\'e}on Bottou, Yoshua Bengio, and Patrick Haffner.
\newblock Gradient-based learning applied to document recognition.
\newblock \emph{Proceedings of the IEEE}, 86\penalty0 (11):\penalty0
  2278--2324, 1998.

\bibitem[Markidis et~al.(2018)Markidis, Der~Chien, Laure, Peng, and
  Vetter]{markidis2018nvidia}
Stefano Markidis, Steven~Wei Der~Chien, Erwin Laure, Ivy~Bo Peng, and Jeffrey~S
  Vetter.
\newblock {NVIDIA} tensor core programmability, performance \& precision.
\newblock In \emph{Proceedings of the International Parallel and Distributed
  Processing Symposium Workshops (IPDPS)}, pp.\  522--531. IEEE, 2018.

\bibitem[Paszke et~al.(2019)Paszke, Gross, Massa, Lerer, Bradbury, Chanan,
  Killeen, Lin, Gimelshein, Antiga, et~al.]{paszke2019pytorch}
Adam Paszke, Sam Gross, Francisco Massa, Adam Lerer, James Bradbury, Gregory
  Chanan, Trevor Killeen, Zeming Lin, Natalia Gimelshein, Luca Antiga, et~al.
\newblock Pytorch: An imperative style, high-performance deep learning library.
\newblock In \emph{Advances in Neural Information Processing Systems}, pp.\
  8024--8035, 2019.

\bibitem[Prechelt(1998)]{prechelt1998automatic}
Lutz Prechelt.
\newblock Automatic early stopping using cross validation: quantifying the
  criteria.
\newblock \emph{Neural Networks}, 11\penalty0 (4):\penalty0 761--767, 1998.

\bibitem[Schwartz et~al.(2019)Schwartz, Dodge, Smith, and
  Etzioni]{Schwartz2019GreenA}
Roy Schwartz, Jesse Dodge, Noah~A. Smith, and Oren Etzioni.
\newblock {Green AI}.
\newblock \emph{arXiv}, abs/1907.10597, 2019.

\bibitem[Silver et~al.(2018)Silver, Hubert, Schrittwieser, Antonoglou, Lai,
  Guez, Lanctot, Sifre, Kumaran, Graepel, et~al.]{silver2018general}
David Silver, Thomas Hubert, Julian Schrittwieser, Ioannis Antonoglou, Matthew
  Lai, Arthur Guez, Marc Lanctot, Laurent Sifre, Dharshan Kumaran, Thore
  Graepel, et~al.
\newblock A general reinforcement learning algorithm that masters chess, shogi,
  and go through self-play.
\newblock \emph{Science}, 362\penalty0 (6419):\penalty0 1140--1144, 2018.

\bibitem[Strubell et~al.(2019)Strubell, Ganesh, and
  McCallum]{strubell2019energy}
Emma Strubell, Ananya Ganesh, and Andrew McCallum.
\newblock Energy and policy considerations for deep learning in {NLP}.
\newblock In \emph{Proceedings of the 57th Annual Meeting of the Association
  for Computational Linguistics (ACL)}, pp.\  3645--3650. ACL, 2019.

\bibitem[Wang et~al.(2018)Wang, Choi, Brand, Chen, and
  Gopalakrishnan]{wang2018training}
Naigang Wang, Jungwook Choi, Daniel Brand, Chia-Yu Chen, and Kailash
  Gopalakrishnan.
\newblock Training deep neural networks with 8-bit floating point numbers.
\newblock In \emph{Advances in Neural Information Processing Systems
  (NeurIPS)}, pp.\  7675--7684, 2018.

\bibitem[Woodland(1989)]{woodland1989weight}
PC~Woodland.
\newblock Weight limiting, weight quantisation and generalisation in
  multi-layer perceptrons.
\newblock In \emph{First IEE International Conference on Artificial Neural
  Networks, (Conf. Publ. No. 313)}, pp.\  297--300. IET, 1989.

\end{thebibliography}
